\documentclass[10pt,twocolumn,letterpaper]{article}

\usepackage[utf8]{inputenc}
\usepackage{iccv}
\usepackage{times}
\usepackage{epsfig}
\usepackage{graphicx}
\usepackage{amsmath}
\usepackage{amssymb}
\usepackage{subfig}
\usepackage{url}

\usepackage[breaklinks=true,bookmarks=false]{hyperref}

\iccvfinalcopy 

\ificcvfinal\fi

\begin{document}

\title{Weighted boxes fusion: Ensembling boxes from different object detection models}

\author{Roman Solovyev\\
Institute for Design Problems in Microelectronics of Russian Academy of Sciences\\
3, Sovetskaya Street, Moscow 124365, Russian Federation\\
{\tt\small roman.solovyev.zf@gmail.com}
\and
Weimin Wang\\
National University of Singapore\\
21 Lower Kent Ridge Rd, Singapore 119077\\
{\tt\small wangweimin777@gmail.com}
\and
Tatiana Gabruseva\\
Cork University Hospital\\
Cork, Ireland\\
{\tt\small tatigabru@gmail.com}
}

\maketitle
\ificcvfinal\thispagestyle{empty}\fi

\begin{abstract}
Object detection is a crucial task in computer vision systems with a wide range of applications in autonomous driving, medical imaging, retail, security, face recognition, robotics, and others. Nowadays, the neural networks-based models are used to localize and classify instances of objects of particular classes. When real-time inference is not required, the ensembles of models help to achieve better results.

In this work, we present a novel method for combining predictions of object detection models: weighted boxes fusion. Our algorithm utilizes confidence scores of all proposed bounding boxes to constructs the averaged boxes.

We tested method on several datasets and evaluated it in the context of the Open Images and COCO Object Detection tracks, achieving top results in these challenges. The 3D version of boxes fusion was successfully applied by the winning teams of Waymo Open Dataset and Lyft 3D Object Detection for Autonomous Vehicles challenges. The source code is publicly available at \url{https://github.com/ZFTurbo/Weighted-Boxes-Fusion}.
\end{abstract}

\section{Introduction}
Object detection is a computer vision technology that deals with detecting instances of semantic objects of a particular class in images and videos~\cite{Szeliski_2010}. Detection is an essential task for a range of practical applications, including autonomous driving~\cite{Geiger2013,Dollar2009}, medical imaging \cite{chexnet_2017,rsna}, robotics, security, and others. The task combines localization with classification. The object detection models typically return the proposed locations of the objects of a given class, class labels, and confidence scores.

The predicted boxes are selected using a non-maximum suppression (NMS) method~\cite{nms}. First, it sorts all detection boxes by their confidence scores. Then, the detection box with the maximum confidence score is selected. At the same time, all other detection boxes with significant overlap to that box are filtered out. It relies on a hard-coded threshold to discard redundant boxes. Some recent works used a differentiable model to learn NMS and introduced soft-NMS~\cite{bodla2017soft} to improve filtering performance.

Such methods work well on a single model. However, they only select the boxes and can not produce averaged localization of predictions combined from various models effectively. Ensembles of models are widely used in applications that do not require real-time inference. Combining predictions from different models generalizes better and usually yields more accurate results compared to a single model ~\cite{okun2011ensembles}. The ensemble methods often get top places in machine learning competitions, for example, see ~\cite{KaggleWin2,rsna,solovyev2020deep,solovyev2020roof,wang2020method,atwood2020inclusive}.

The other technique, commonly-used in practice, is a test-time augmentation (TTA). In this method, the predictions of the same model are obtained for the original and augmented images (for example, vertically/horizontally reflected) and then averaged.

In this paper, we propose a novel Weighted Boxes Fusion (WBF) method for combining predictions of object detection models.
Unlike NMS and soft-NMS methods that simply remove part of the predictions, the proposed WBF method uses confidence scores of all proposed bounding boxes to constructs the average boxes.

This method significantly improves the quality of the combined predicted rectangles. The technique was evaluated in the context of the Open Images Object Detection Challenge and helped achieving one of the top results in the challenge~\cite{kaggle_result}. The source code is publicly available at \url{https://github.com/ZFTurbo/Weighted-Boxes-Fusion}.

The method can also be beneficial for the ensemble of the manual labels in the medical applications, i.e., combining labels of different expert radiologists in the pneumonia detection task. The averaged predictions of the expert panel should lead to more accurate labels. Hence it can produce better data for the computer-assisted diagnostics programs.

\section{Related Work}

\subsection{Non-maximum suppression (NMS)}
Predictions of a model in the image detection task consist of coordinates of the bounding box rectangle, a class label of the object, and a confidence score (probability from 0 to 1) that reflects how confident the model is in this prediction.

\begin{figure}[ht!]
    \centering
    \includegraphics[width=1\linewidth]{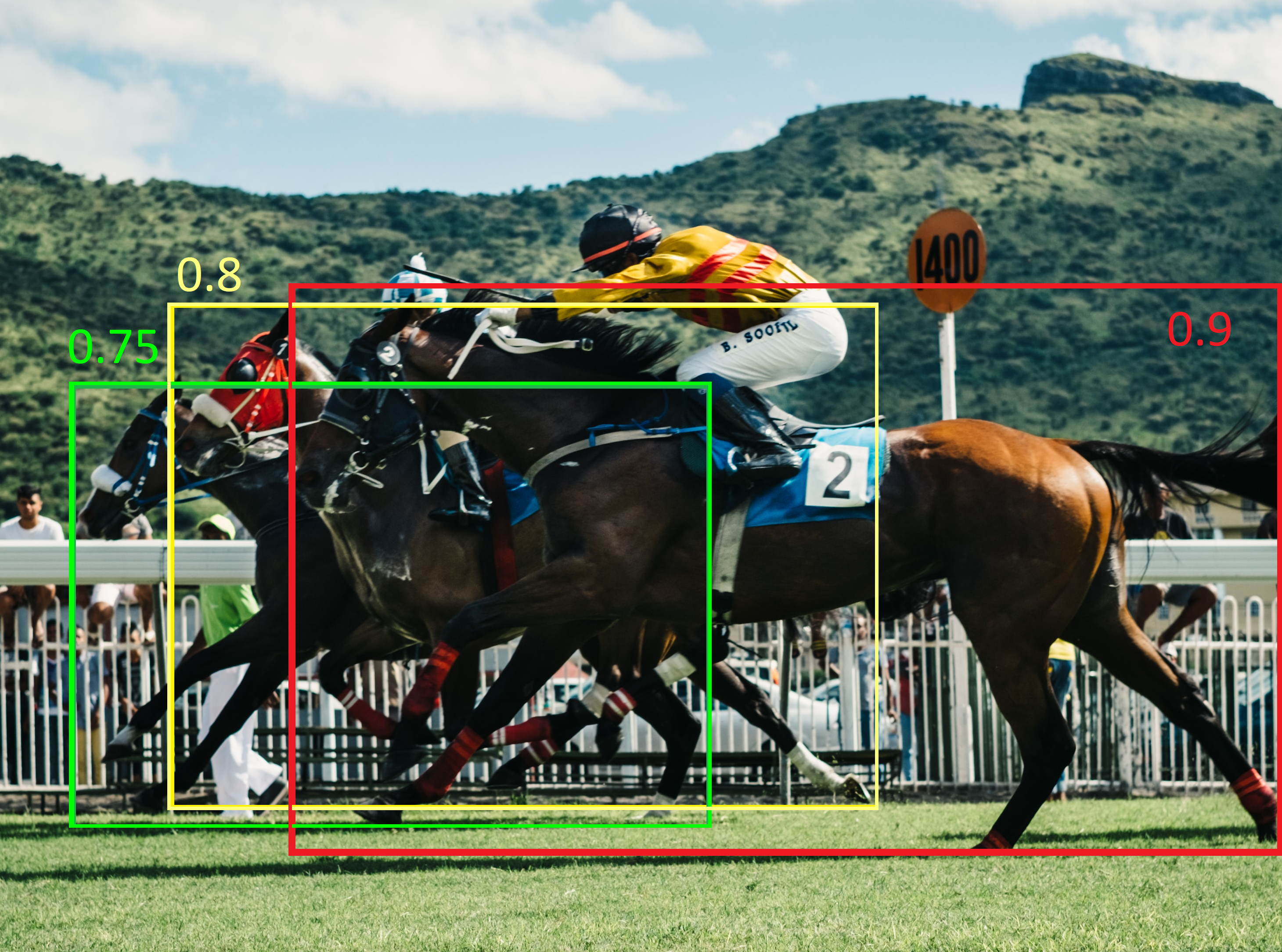}
    \caption{The photo illustrates several overlapping horses in the race. For several detections with the high confidence score, only one will be selected by the NMS algorithm for the IoU threshold above $0.5$. Photo by Julia Joppien on Unsplash.}
    \label{fig:horses}
\end{figure}

In the NMS method, the boxes are considered as belonging to a single object if their overlap, intersection-over-union (IoU) is higher than some threshold value. Thus, the boxes filtering process depends on selection of this single IoU threshold value, which affects the performance of the model. However, setting this threshold is tricky: if there are objects side by side, one of them would be eliminated. Figure~\ref{fig:horses} shows an example illustrating one such case. For the IoU threshold of $0.5$, only one box prediction will remain. The other overlapping objects detected will be removed. Such errors reduce the precision of the model.

\subsection{Soft-NMS}
The soft-NMS method was proposed in~\cite{bodla2017soft} to reduce this problem. Instead of completely removing the detection proposals with high IoU and high confidence, it reduces the confidences of the proposals proportional to IoU value. On the example above, soft-NMS will lower the confidence scores proportionally to the IoU overlap. As the green detection box has a significant overlap with the yellow one, it will be removed. The recent adversarial attack has exploited this problem of the NMS and soft-NMS methods~\cite{Wang20advers}.

Soft-NMS demonstrated noticeable improvements over traditional NMS on standard benchmark datasets, like PASCAL VOC and MS COCO~\cite{cocodataset}. It has shown to improve performance for single models~\cite{bodla2017soft}. However, both NMS and soft-NMS discard redundant boxes, and thus can not produce averaged localization predictions from different models effectively.

\section{Weighted Boxes Fusion}
Here, we describe the novel method for the boxes fusion: Weighted Boxes Fusion (WBF). Suppose, we have bounding boxes predictions for the same image from $\mathbf{N}$ different models. Alternatively, we have $\mathbf{N}$ predictions of the same model for the original and augmented versions of the same image (i.e., vertically/horizontally reflected).

The WBF algorithm works in the following steps:
\begin{enumerate}
\item Each predicted box from each model is added to a single list $\mathbf{B}$. The list is sorted in decreasing order of the confidence scores $\mathbf{C}$.

\item Declare empty lists $\mathbf{L}$ and $\mathbf{F}$ for boxes clusters and fused boxes, respectively. Each position in the list $\mathbf{L}$ can contain a set of boxes (or single box), which form a cluster; each position in $\mathbf{F}$ contains only one box, which is the fused box from the corresponding cluster in $\mathbf{L}$. The equation to generate fused boxes will be discussed later.

\item Iterate through predicted boxes in $\mathbf{B}$ in a cycle and try to find a matching box in the list $\mathbf{F}$. The match is defined as a box with a large overlap with the box under question ($IoU > \mathbf{THR}$). Note: in our experiments, $\mathbf{THR}=0.55$ was close to an optimal threshold.

\item If the match is not found, add the box from the list $\mathbf{B}$ to the end of lists $\mathbf{L}$ and $\mathbf{F}$ as new entries; proceed to the next box in the list $\mathbf{B}$.

\item If the match is found, add this box to the list $\mathbf{L}$ at the position $\mathbf{pos}$ corresponding to the matching box in the list $\mathbf{F}$.

\item Recalculate the box coordinates and confidence score in $\mathbf{F[pos]}$, using all $\mathbf{T}$ boxes accumulated in cluster $\mathbf{L[pos]}$, with the following fusion formulas:

\begin{equation}
    \mathbf{C} = \frac{\sum_{i=1}^{\mathbf{T}} \mathbf{C}_i}{T},
\end{equation}

\begin{equation}
    \mathbf{X1,2} = \frac{\sum_{i=1}^{\mathbf{T}} \mathbf{C}_i * \mathbf{X1,2}_i}{\sum_{i=1}^{\mathbf{T}} \mathbf{C}_i},
\end{equation}

\begin{equation}
    \mathbf{Y1,2} = \frac{\sum_{i=1}^{\mathbf{T}} \mathbf{C}_i * \mathbf{Y1,2}_i}{\sum_{i=1}^{\mathbf{T}} \mathbf{C}_i},
\end{equation}
Note: we can use some nonlinear weights as well, for instance, $\mathbf{C^2}$, $sqrt(\mathbf{C})$, etc.

Set the confidence score for the fused box as the average confidence of all boxes that form it.
Coordinates of the fused box are weighted sums of the coordinates of the boxes that form it, where the weights are confidence scores for the corresponding boxes. Thus, boxes with larger confidence contribute more to the fused box coordinates than boxes with lower confidence.

\item After all boxes in $\mathbf{B}$ are processed, re-scale confidence scores in $\mathbf{F}$ list: multiply it by a number of boxes in a cluster and divide by a number of models $\mathbf{N}$. If a number of boxes in the cluster is low, it could mean that only a small number of models predict it. Thus, we need to decrease confidence scores for such cases.
It can be done in two ways:
\begin{equation}
    \mathbf{C} = \mathbf{C} * \frac{min(T, N)}{N},
\end{equation}
or
\begin{equation}
    \mathbf{C} = \mathbf{C} * \frac{T}{N},
\end{equation}

In practice, the outcome for both variants did not differ significantly, with the first being slightly better. In some cases, a number of boxes in one cluster can be more than a number of models.
\end{enumerate}

Both NMS and soft-NMS exclude some boxes, while WBF uses all boxes. Thus, it can fix cases where all boxes are predicted inaccurately by all models. This case is illustrated in Figure\ref{fig:ctr-vis}. NMS/soft-NMS will leave only one inaccurate box, while WBF will fuse it using all predicted boxes.
\begin{figure}[ht!]
    \centering
    \includegraphics[width=1\linewidth]{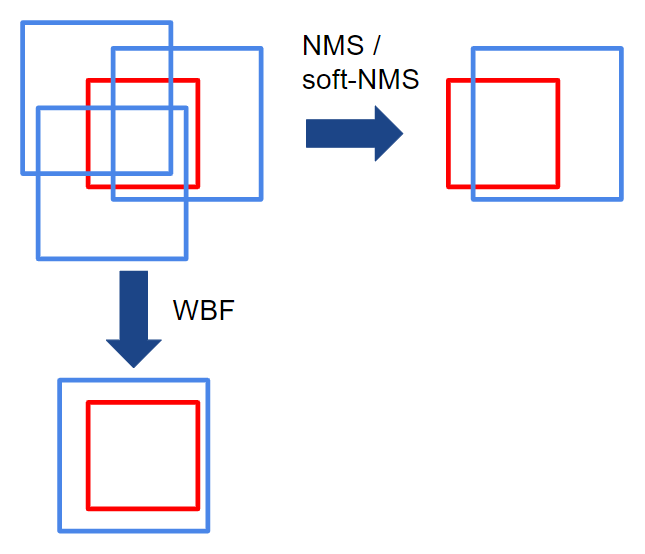}
    \caption{Schematic illustration of NMS/soft-NMS vs. WBF outcomes for an ensemble of inaccurate predictions. Blue – different models' predictions, red – ground truth.}
    \label{fig:ctr-vis}
\end{figure}

The non-maximum weighted (NMW) method proposed in~\cite{zhou2017cad,Ning2017} has a similar idea. However, the NMW method does not change confidence scores; it uses IoU value to weight the boxes. NMW uses a box with the highest confidence to compare with, while WBF updates a fused box at each step, and uses it to check the overlap with the next predicted boxes. Also, NMW does not use information about how many models predict a given box in a cluster and, therefore, does not produce the best results for models' ensemble. We compare our proposed WBF method with NMW, NMS, and soft-NMS techniques in section 6.

\section{Datasets}
In this section, we describe data sets used to test the proposed WBF method of boxes fusion.
We used the Open Images data set~\cite{kuznetsova2018open}, the largest data set with object detection annotations to date, and Microsoft COCO~\cite{cocodataset,Lin2014}, one of the most popular benchmark datasets for object detection.

\subsection{Open Images Dataset}
The Open Images Object Detection Challenges held at the International Conference on Computer Vision 2019 and hosted on Kaggle~\cite{kaggle_overview}. The challenge used V$5$ release of the Open Images dataset~\cite{kuznetsova2018open} that includes around $16$M bounding boxes for $600$ object classes on $1.9$M images. To date, it is the largest labeled dataset with object detection annotations.

In the Object Detection track of the challenge, the participants needed to predict a tight bounding box around object instances. The training set contained $12.2$M bounding-boxes across $500$ categories on $1.7$M images. The boxes have mainly been manually drawn by professional annotators to ensure accuracy and consistency. The images of the dataset are very diverse and often contain complex scenes with several objects (7 per image on average). The details on the Open Images dataset are in~\cite{kuznetsova2018open}.

The dataset consists of the training set ($1743042$ images), validation set ($41620$ images), and the test set ($99999$ images). The validation and test sets, as well as part of the training set, have human-verified image-level labels. As the test set labels are not available, we performed the ablation study on the validation set. The test set results were accessible through submissions on the Kaggle web site.

\subsection{COCO Dataset}
The Microsoft COCO dataset became a popular benchmark for image detection and segmentation tasks~\cite{Lin2014}. The dataset contains over $200,000$ images with $80$ object categories. Annotations for the training and validation sets (with over $500,000$ object instances) are publicly available here~\cite{cocodataset}.
The models used in the ablation study presented in this paper were trained on the train2017 with $118$k labeled images. Then, we tested the models on COCO validation set val2017 with $5$K images. The best results were also submitted to the official leader board and evaluated on the test set with $20$k images.~\cite{cocoleaderboard}.

\section{Evaluation}
For the Open Images Object Detection challenge, the evaluation metric was chosen by organizers. The models were evaluated using the mean average precision (mAP) at intersection-over-union (IoU) value equal to $0.5$.

The IoU is a ratio of overlap between two objects (A and B) to the total area of the two objects combined. The IoU of a set of predicted bounding boxes (A) and ground truth bounding boxes (B) is calculated as:
\begin{equation}
    IoU(A,B) = \frac{A \cap B}{ A \cup B}
\end{equation}

At each threshold value $t$, a precision value is calculated from the numbers of true positives (TP), false negatives (FN), and false positives (FP) as:
\begin{equation}
    Precision(t) = \frac{TP(t)}{TP(t) + FP(t) + FN(t)}
\end{equation}

For the Open Images Object Detection challenge, the AP is calculated at IoU = $0.5$. The final $mAP$ is computed as the average AP over the $500$ classes in the challenge. The metric calculation is described on the Open Images Challenge website~\cite{evaluation}. The implementation of this $mAP$ variant is publicly available as part of the Tensorflow Object Detection API~\cite{tf_od_api} under the name 'OID Challenge Object Detection Metric'.

For the COCO dataset, the mAP was computed for different intersection-over-union (IoU) thresholds. The metric sweeps over a range of IoU thresholds, at each point calculating an average precision value. The threshold values range from $0.5$ to $0.95$ with a step size of $0.05$: (0.5, 0.55, 0.6, 0.65, 0.7, 0.75, 0.8, 0.85, 0.9, 0.95). A predicted object is considered a "hit" if its intersection over union with a ground truth object is greater than $0.5$.

The average precision ($AP$) of a single image is calculated as the mean of the above precision values at each IoU threshold:
\begin{equation}
    AP = \frac{1}{|thresholds|} \sum_t \frac{TP(t)}{TP(t) + FP(t) + FN(t)}
\end{equation}

The python implementation of this $mAP$ metric variant can be found, for example, in~\cite{bes_mAP_python}.

\section{Experiments}
In this section, we perform an ablation study and compare the results obtained from WBF with NMS, soft-NMS, and the weighted NMW method. We also conduct specific experiments to understand when WBF gives the best boost to model performance. We used different datasets and models to make experiments more representative.

\subsection{Models}
A set of different detector models was used for the ablation study, including single-stage and two-stage detectors.
Among single-shot detectors, we used RetinaNet \cite{lin2017focal} implementation in Keras with different ResNet~\cite{resnet} backbones, pre-trained on the Open Images Dataset~\cite{kuznetsova2018open}. For the COCO dataset, we used different versions of EfficientDet models~\cite{Tan2019} from \cite{effdet2} and DetectoRS model~\cite{Qiao2020} from the official repository~\cite{detectoRS} with weights trained on MS COCO. All individual models produced predictions selected using the NMS algorithm.

The two-stage detectors used included Faster R-CNN~\cite{ren2015faster}, Mask R-CNN~\cite{He2017}, and Cascade R-CNN~\cite{cai2018cascade}.

\subsection{An ensemble of two different models}

\begin{table*}[htb!]
\centering
\begin{tabular}{l c c c}
\hline
Model & mAP(0.5..0.95) & mAP(@0.5 IoU) & mAP(@0.5 IoU) \\
\hline
EffDetB6 & 0.513 & 0.705 & 0.554 \\
EffDetB7 & 0.521	& 0.710 & 0.562\\
\hline
Method & mAP(0.5..0.95) & mAP(0.5) & mAP(0.75) \\
\hline
NMS & 0.5269 & 0.7156 & 0.5737\\
IoU threshold & 0.76 & 0.51 & 0.85 \\
Weights & [3, 4] & [3, 4] & [2, 3] \\
\hline
Soft-NMS & 0.5239 & 0.7121 & 0.5677\\
Threshold & 0.0014 & 0.0013 & 0.0017 \\
Sigma & 0.15 & 0.145 & 0.16 \\
Weights & [3, 4] &  [3, 4] & [3, 4] \\
\hline
NMW & 0.5285 & 0.7171 & 0.5743 \\
IoU threshold& 0.80 & 0.50 & 0.83 \\
Weights & [3, 4] & [1, 1] & [3, 4] \\
\hline
WBF & \textbf{0.5344} & \textbf{0.7244} & \textbf{0.5824} \\
IoU threshold & 0.75 & 0.59 & 0.77 \\
Weights & [2, 3] & [3, 4] & [3, 4] \\
\hline
\end{tabular}
\caption{Calculated mAP for individual EfficientDet models and ensembles of their predictions. The boxes were combined using the NMS, soft-NMS, NMW, and WBF methods. The optimal parameters are listed below for each method.}
\label{table:coco}
\end{table*}

\begin{table*}[htb!]
\centering
\begin{tabular}{l c c c}
\hline
Model & mAP(0.5..0.95) & mAP(@0.5 IoU) & mAP(@0.5 IoU) \\
\hline
EffDetB7 & 0.521 & 0.710 & 0.562 \\
EffDetB7 (Mirror) & 0.519	& 0.710 & 0.558\\
\hline
Method & mAP(0.5..0.95) & mAP(0.5) & mAP(0.75) \\
\hline
NMS & 0.5233 & 0.7129 & 0.5656 \\
\hline
Soft-NMS & 0.5210 & 0.7092 & 0.5633 \\
\hline
NMW & 0.5250 & 0.7138 & 0.5691 \\
\hline
WBF & \textbf{0.5262} & \textbf{0.7144} & \textbf{0.5717} \\
\hline
\end{tabular}
\caption{Calculated mAP for TTA ensemble of predictions from the EfficientDetB7 model. We combine predictions for original and horizontally mirrored images.}
\label{table:cocotta}
\end{table*}

Table 1 shows results for MS COCO validation set for two models (EfficientDetB6 and EfficientDetB7 \cite{effdet2} with NMS outputs) and their predictions ensemble. These models have recently shown top results on the COCO database \cite{Tan2019}. The predictions were combined using four different techniques: NMS, soft-NMS, NMW, and WBF. We compare the resulting mAP(0.5...0.95), mAP(0.5) and mAP(0.75).

We used a grid search to find the optimal parameters for each of the four methods, which allowed achieving the best $mAP$ for the given method. The individual models had different results on the MS COCO data set. Therefore, we also varied the weights for the model predictions. The optimal parameters are listed in Table 1. Even for the ensemble of only two models, the WBF method outperformed significantly other methods.

\subsection{Test-time-augmentation ensemble}

We tested the performance of different boxes combining methods for the TTA ensemble. We used the EfficientNetB7 model trained on the COCO dataset. We predict boxes for both original and horizontally mirrored images. Then, we perform a grid search to find optimal parameters for all four ensemble methods. In this case, we used identical weights for both predictions. The results are in Table 2.

\subsection{An ensemble of many different models}

\subsubsection{Ensemble of models for COCO Dataset}

\begin{figure*}[htb!]
    \centering
    \includegraphics[width=1\linewidth]{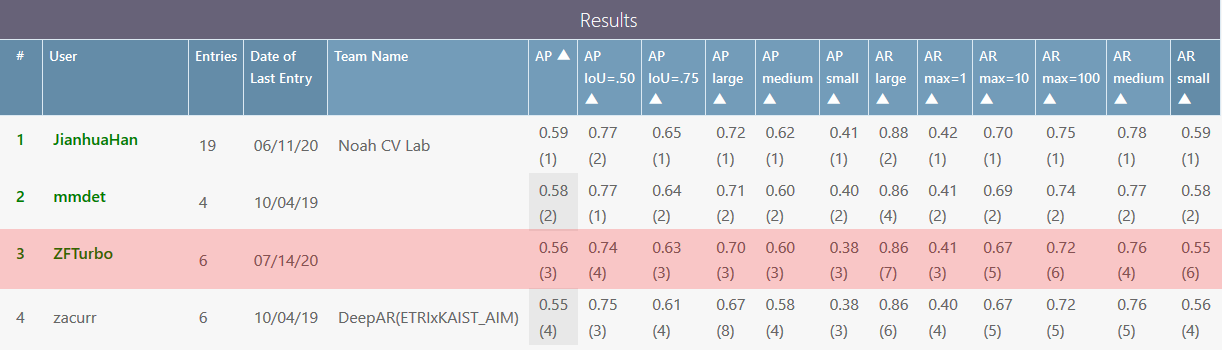}
    \caption{The official leader boards of COCO Detection Challenge (Bounding Box) as of 07/08/20. The third top entry represents the results described in this paper. These results can be reproduced using benchmark published on GitHub~\cite{benchmarkCocoWBF}.}
    \label{fig:ctr-vis}
\end{figure*}
Here, we combined the results of several different models. We used several models trained on MS COCO models from the official EfficientDet repository \cite{effdet2}, two DetectoRS~\cite{Qiao2020,detectoRS} models with two different backbones, and the YOLOv5 model~\cite{yolov5}. For this experiment, we performed TTA and made predictions for both original and horizontally flipped images. The individual models' performance and the result for their ensemble are in Table 3. Each model used the NMS method for its output boxes prediction.

The weights in the final ensemble and IoU threshold for the WBF methods were optimized using the validation set. The WBF method for combining predictions of detection models gave \textbf{56.1} mAP for the validation data set and \textbf{56.4} mAP for the test set. It is a considerable improvement compared to the performance of individual models and gives the top third result on the MS COCO official leaderboard \cite{cocoleaderboard} (as of 07/08/2020). These results can be reproduced using a benchmark published on GitHub~\cite{benchmarkCocoWBF}.

\subsubsection{Ensemble of RetinaNet models for Open Images Dataset}

\begin{table*}[htb!]
\centering
\begin{tabular}{l c c}
\hline
 Model & mAP @(0.5..0.95) & mAP @(0.5..0.95)  \\
  & original images & flipped images \\
 \hline
EfficientNetB4 & 49.0 &	48.8   \\	
EfficientNetB5 & 50.5 &	50.2	\\
EfficientNetB6 & 51.3 &	51.1	\\
EfficientNetB7 & 52.1 &	51.9	\\
DetectoRS (X101) & 51.5 &	51.5	\\
DetectoRS (ResNet50) & 49.6 &	49.6	\\
YOLO v5x (TTA) & 50.0 &	 -	\\
\hline
Ensemble & COCO validation & COCO test \\
\hline
WBF & \textbf{56.1} & \textbf{56.4} \\
IoU threshold & 0.7 & 0.7 \\
Weigths & [4,5,7,9,8,5,5] & [4,5,7,9,8,5,5] \\
\hline
\end{tabular}
\caption{The mAP metrics for several individual models obtained on MS COCO validation set for the original and horizontally flipped images, and the resulting mAP metrics for the ensemble of their predictions.}
\label{table:3}
\end{table*}

\begin{table}[htb!]
\centering
\begin{tabular}{l c}
\hline
 RetinaNet backbone & mAP (@0.5 IoU) \\
 \hline
 ResNet152 & 0.5180 \\
 ResNet101 & 0.4997 \\
 ResNet50 & 0.4613 \\
 ResNet152 (v2) & 0.5123 \\
 \hline
Method & mAP (@0.5 IoU) \\
\hline
NMS & 0.5442 \\
IoU threshold& 0.3 \\
\hline
Soft-NMS & 0.5428\\
Sigma & 0.05 \\
\hline
WBF & \textbf{0.5665} \\
IoU threshold & 0.55\\
\hline
\end{tabular}
\caption{The mAP metrics for individual RetinaNet models and ensembles of their predictions. We used the same weights for each model.}
\label{table:3}
\end{table}

The results obtained on the Open Images Dataset~\cite{kuznetsova2018open} for combining predictions of RetinaNet single-shot-detector models with different backbones are in Table 4. For the Open Images challenge, we used mAP at $0.5$ IoU, suggested by the challenge organizers. Again, we used a grid search to find the optimal parameters for each of the methods. Combining predictions from several detectors with different backbones yields better performance, and the WBF method demonstrated the best results.

\subsubsection{Ensemble of fairly different models for Open Images Dataset}

\begin{table*}[htb!]
\centering
\begin{tabular}{l c c}
 \hline
 Model & Backbone & mAP (@0.5 IoU) \\
 \hline
 RetinaNet (Mix) & resnet50, resnet101, resnet152 & 0.5164 \\
 Mask-RCNN & ResNext101 & 0.5019 \\
 Cascade-RCNN & ResNext101-FPN-GroupNorm & 0.5144 \\
 MMDetection HTC & X-101-64x4d-FPN & 0.5152 \\
 Faster RCNN & InceptionResNetV2 & 0.4910 \\
\hline
Method & Params & mAP (@0.5 IoU) \\
\hline
 NMS & IoU threshold = 0.5 & 0.5642 \\
 Soft-NMS & Sigma 0.1, threshold 1e-03 & 0.5616 \\
 NMW &IoU threshold 0.5 & 0.5667\\
 WBF & IoU threshold 0.6 & \textbf{0.5982}\\
\end{tabular}
\caption{Object detection models trained on Open Images dataset, the resulted mAPs for these models and for their ensembles obtained via NMS, soft-NMS, NMW, and WBF methods.}
\label{table:1}
\end{table*}

In this experiment, performed on the Open Images dataset, we used a range of different models. The ensemble included the RetinaNet models combination described above, Mask-RCNN with ResNext101 backbone~\cite{tensorFasterRCNN2019}, Cascade-RCNN with ResNext101 backbone, FPN, and GroupNorm~\cite{tensorFasterRCNN2019}, MMdetection HTC model~\cite{MMDetHTC2019}, and Faster-RCNNs model with InceptionResNetV2 backbone from~\cite{TensorflowDetectionModelZoo2019}. These models had a comparable performance on the Open Images dataset in terms of mAP. In the previous experiments, we studied the performance of the WBF method for similar models. Here, we explore combining predictions from highly different models.

The WBF method outperformed by far the other algorithms (see Table 5) and helped to achieve one of the top results (7 out of 558) in the challenge~\cite{kaggle_result}.

\section{Discussion}
Most neural networks for Object Detection use NMS or Soft-NMS output to filter predicted boxes. We tested our method for models' predictions that have already been processed by NMS. In this experiment, we wanted to check whether it is possible to use WBF at the output of a model instead of NMS. For the experiment, we used the RetinaNet~\cite{lin2017focal} detector with the ResNet152~\cite{resnet} backbone trained on the Open Images dataset. After we disabled the last NMS layer, we got $500$ boxes with their probability scores. We compared the mAP (@0.5 IoU) metric for combining these output predictions via NMS and WBF algorithms.

The results are the following:
\begin{enumerate}
\item Raw boxes (no NMS/WBF) - mAP: 0.1718 - the value is rather small because of many overlapping boxes
\item NMS with default IoU threshold = 0.5 (e.g. standard model output) -- mAP: 0.4902
\item NMS with optimal IoU threshold = 0.47 -- mAP: 0.4906 -- the tiny change from the default threshold
\item WBF with optimal parameters -- mAP: 0.4532 (the optimized parameters: IoU threshold = 0.43, skip threshold = 0.21)
\end{enumerate}
As can be seen from the experiment, replacing the NMS method at the output of the individual model with WBF results in the degradation of the model performance.
We think it is due to the excessive number of low scored wrong predictions given by the model output. The NMS or soft-NMS produces more efficient filtering for those predictions, while the WBF works better when used for the models' ensemble.

Thus, WBF works well for combining boxes for fairly accurate models. However, when it comes to a large number of overlapping boxes with different confidence scores, WBF gives worse results than NMS.

The method was expanded on the 3D boxes fusion as well. The code for the 3D boxes fusion is available here:~\cite{wbf3d}. In the recent Waymo Open Dataset Challenge~\cite{waymo_challenge}, the winners from the first place of 3D Detection and Domain Adaptation track~\cite{Ding2020a} and the second place from the 2D Object Detection track~\cite{Chen2020} used this method for combining predictions. The first place winners from Lyft 3D Object Detection for Autonomous Vehicles~\cite{KaggleLyft} reported that WBF gives better score for their ensemble (0.222 vs 0.220)~\cite{KaggleWin3}. It was also reported that the WBF method gives the best results on the Wheat detection dataset for boxes ensemble comparing to other methods~\cite{WheatOpt}.

WBF is currently slower than the NMS and soft-NMS methods. While the speed depends on the set of boxes, a number of classes, and software implementation, the WBF algorithm, on average, is around three times slower than the standard NMS. 

\section{Conclusion}
In this work, we proposed a new technique for combining predictions of object detection models, both for 2D and 3D boxes. The described WBF method uses confidence scores of all proposed bounding boxes in the iterative algorithm that constructs the averaged boxes. We evaluated our method and compared its performance to the alternatives on two popular benchmark datasets for object detection: the Open Images~\cite{kuznetsova2018open} and Microsoft COCO~\cite{Lin2014} datasets.

The WBF method outperformed by far other results for combining predictions. It helped achieving one of the top results in the Open Images Detection Challenge and the top performance for the COCO Detection Challenge ((\textbf{56.1} mAP for validation data set and \textbf{56.4} mAP for the test-dev set). It's a considerable improvement compared to the performance of individual models.

The method is also available for the 3D boxes fusion. The 3D version of the WBF algorithm was successfully applied by the winners of the Waymo Detection challenge \cite{Ding2020a,Chen2020} and Lyft 3D Object Detection for Autonomous Vehicles challenge~\cite{KaggleLyft,KaggleWin3}. The source code for the WBF and the usage examples is available at GitHub: \url{https://github.com/ZFTurbo/Weighted-Boxes-Fusion}.

{\small
\bibliographystyle{ieee_fullname}
\bibliography{tanya}
}

\end{document}